\pdfoutput=1

\documentclass[11pt]{article}

\usepackage[final]{acl}

\usepackage{times}
\usepackage{latexsym}
\usepackage{enumitem}

\usepackage[T1]{fontenc}

\usepackage[utf8]{inputenc}

\usepackage{microtype}

\usepackage{inconsolata}

\usepackage{graphicx}
\usepackage{tabularray}

\usepackage{amsmath}
\usepackage{amssymb}

\usepackage[utf8]{inputenc}
\usepackage[T1]{fontenc}
\usepackage{tcolorbox}
\usepackage{xcolor}

\usepackage{lipsum}
\usepackage{microtype}
\title{Enhancing Training Data Attribution for Large Language Models with Fitting Error Consideration}

\author{Kangxi Wu$^{1,2}$, Liang Pang$^{1}$\thanks{\ \ Corresponding author},  Huawei Shen$^{1,2}$, Xueqi Cheng$^{1,2}$ \\
  $^{1}$ Key Laboratory of AI Safety, Chinese Academy of Sciences\\  Institute of Computing Technology, CAS\\
  $^{2}$ University of Chinese Academy of Sciences \\
  \texttt{\{wukangxi22s, pangliang, shenhuawei, cxq\}@ict.ac.cn}
}

\begin{document}
\maketitle
\begin{abstract}
The black-box nature of large language models (LLMs) poses challenges in interpreting results, impacting issues such as data intellectual property protection and hallucination tracing. Training data attribution (TDA) methods are considered effective solutions to address these challenges.
Most recent TDA methods rely on influence functions, assuming the model achieves minimized empirical risk. However, achieving this criterion is difficult, and sourcing accuracy can be compromised by fitting errors during model training. In this paper, we introduce a novel TDA method called Debias and Denoise Attribution (DDA), which enhances influence functions by addressing fitting errors. Specifically, the debias strategy seeks to improve the performance of influence functions by eliminating the knowledge bias present in the base model before fine-tuning, while the denoise strategy aims to reduce discrepancies in influence scores arising from varying degrees of fitting during the training process through smoothing techniques.
Experimental results demonstrate that our method significantly outperforms existing approaches, achieving an averaged AUC of 91.64\%. Moreover, DDA exhibits strong generality and scalability across various sources and different-scale models like LLaMA2, QWEN2, and Mistral.~\footnote{Our code: \url{https://github.com/cansee5/DDA}} 

\end{abstract}

\section{Introduction}
Large language models (LLMs) represent a novel paradigm of pre-training followed by fine-tuning, which is premised on initially training a model with a massive number of parameters on an extensive corpus to establish a foundational model~\cite{zhao2023survey}. 
LLMs achieve a significant breakthrough in the scale of training data and the quantity of model parameters~\cite{kaplan2020scaling}, thereby substantially enhancing the overall capability of model in executing complex tasks~\cite{touvron2023llama, openai2024gpt4, geminiteam2024gemini}.
However, LLMs still face a series of challenges and issues in practical applications. The complexity, black-box nature, and randomness of the training process of these models make it difficult to accurately understand and control their behavior and output~\cite{ding2023maclasa, zhao2024explainability}. 
This presents significant challenges for the interpretability of LLMs, the protection of intellectual property rights associated with data, and the tracking of model-generated hallucinations, thereby posing potential risks to the deployment and use of LLMs~\cite{xu2023ai, cui2024risk, dai2024unifying}.

The numerous challenges and issues faced by LLMs in practical applications have garnered widespread attention from the research community~\cite{kaddour2023challenges}. To ensure their lawful, fair, and trustworthy use, researchers are continuously exploring methods to address these problems~\cite{YAO2024100211, liu2023trustworthy}. Previous work has partially explained the model outputs from the perspective of training data, including methods such as influence functions~\cite{koh2017understanding, akyurek-etal-2022-towards, ladhak2022contrastive, Park2023trak, grosse2023studying} and watermark embedding~\cite{wang2023wasa, hu2024unbiased}. 
However, the watermark embedding methods assume an overly idealized scenario, requiring specific interventions during the training phase, which significantly limits its applicability~\cite{sadasivan2023can, pang2024free}. Consequently, for TDA in deep learning models, the methods based on influence functions are preferred due to their robust theoretical foundation, ability to provide high-precision, fine-grained attribution, and the absence of a need for additional modifications to the training data or the model itself, making them applicable across a wide range of scenarios.
Nevertheless, the methods based on influence functions are highly dependent on the accuracy of gradient approximation~\cite{koh2017understanding, bae2022if, nguyen2024bayesian}. Due to the training process of LLMs often failing to meet the criteria for empirical risk minimization (ERM)~\cite{vapnik1999overview, zhu2024crossmodelcomparativelossenhancing}, methods based on TDA utilizing influence functions are easily impacted by fitting errors encountered during model training, resulting in reduced effectiveness in attributing training data.

To address issues of influence functions, we propose a method called Debias and Denoise Attribution (DDA) to correct the influence function by reducing the impact of fitting errors. First, we continue training LLMs from a base model to obtain multiple versions of LLMs with different fitting errors (training losses). In the debias phase, we correct each influence score by subtracting the influence score of the base model. Next, in the denoise phase, we eliminate discrepancies in the sourcing results by averaging the rectified influence scores of all LLM versions. Theoretically, we prove that DDA can align the influence functions more closely with the ERM theoretical conditions, thereby enhancing the effectiveness of influence function-based TDA tasks.

However, evaluating training data attribution tasks is not well-defined and very hard~\cite{wang2023evaluating}, thus we transform it into a hallucination tracing task. Specifically, we construct a hallucination dataset named "Hallucination Xsum" and assess the performance of LLMs in tracing training data under hallucination conditions to measure the effectiveness of the baseline and DDA methods. This transformation approach allows us to more effectively evaluate the accuracy and reliability of training data attribution methods. Experimental results show that DDA significantly outperforms methods such as BM25~\cite{robertson2009probabilistic}, TracIN~\cite{pruthi2020estimating}, TRAK~\cite{Park2023trak}, and CEA~\cite{ladhak2022contrastive} in training data attribution, with an AUC value as high as 93.49\%. Moreover, DDA demonstrates its generality and scalability across models like  LLaMA2~\cite{touvron2023llama}, QWEN2~\cite{qwen}, and Mistral~\cite{jiang2023mistral}.

Our main contributions can be summarized as follows:

\begin{itemize}[leftmargin=*]
    \item We propose a novel approach for TDA by refining influence functions using debias and denoise strategies. This method mitigates the impact of fitting errors during the training process of LLMs on TDA methods based on influence functions, thereby achieving state-of-the-art performance.
    \item Through theoretical derivation, we demonstrate that the debias and denoise strategies contribute to enhancing the accuracy of TDA methods based on influence functions within LLMs, thereby validating the feasibility of applying DDA for conducting TDA in LLMs.
    \item We conduct extensive experiments on
    hallucination datasets. The results demonstrate that our proposed method outperforms all baseline methods, achieving SOTA performance. Meanwhile, we also proved that the DDA method has stronger generalizability and scalability.
\end{itemize}

\section{Related Work}
The existing methods for training data attribution
can be broadly categorized into two types: watermark-based and influence-based attribution.

\subsection{The Watermark-based Data Attribution}
Embedding unique watermark information in the training corpus is a technique used to achieve training data traceability, aiming to ensure that the model can learn and reflect this watermark information. ~\cite{hu2023unbiased} proposed an unbiased watermarking technique that balances content traceability and quality assurance, ensuring that the model output contains watermarks without affecting the performance of model  in downstream tasks. The WASA framework proposed by ~\cite{wang2023wasa} provides an effective solution for content traceability and data provenance by training large language models (LLMs) to learn the mapping between the text from different data providers and their unique watermarks. Additionally, the research of Meta research on the radiology of watermarked texts reveals the detectability of watermarked training data, indicating that even embedding a small amount of watermarked data can make the model output highly recognizable~\cite{sander2024watermarking}.

However, watermarking technology faces several challenges and limitations~\cite{sadasivan2023can}. Firstly, watermarks are vulnerable to attacks or removal, significantly diminishing their effectiveness. Current techniques can effectively remove or alter watermarks while preserving the original data characteristics~\cite{pang2024attacking, wei2024stable, deng2024unke}, thereby compromising their content traceability. Moreover, applying watermarking frameworks to pre-trained models is challenging because they require specific interventions during the training process~\cite{kirchenbauer2024on}. This requirement limits the applicability of watermarking technology to existing models, especially when training data or the training process is inaccessible. Lastly, embedding watermarks may adversely affect the quality of generated text, as the process can introduce noise or other disturbances, reducing the naturalness and readability of the text~\cite{piet2023mark}.

\subsection{The Influence-based Data Attribution}
Understanding and quantifying the influence of each training sample on the predictions of a machine learning model is crucial for diagnosing and improving model performance. Influence functions~\cite{hampel1974influence}, as a primary method for this purpose, help identify the most critical training data points that affect model decisions. They are particularly useful for detecting harmful samples or outliers in the training data. However, calculating the inverse of the Hessian matrix, which is required for influence functions~\cite{koh2017understanding}, becomes computationally expensive and complex when dealing with a large number of parameters.
To address these challenges, ~\cite{pruthi2020estimating} proposes TracIN, a method that approximates influence using first-order gradients and checkpoints. This approach significantly reduces computational complexity and scales effectively to large datasets and complex models.

In the context of NLP tasks, ~\cite{pezeshkpour2021empirical} analyzes gradient-based methods (such as influence functions and their variants) and simple similarity-based retrieval methods in terms of their efficiency and complexity when applied to BERT models. They found that gradient-based methods, while theoretically robust, often suffer from high computational costs. To mitigate this, ~\cite{akyurek-etal-2022-towards} adopted a common re-ranking strategy from information retrieval. Instead of applying training data attribution methods to the entire training set, they first create a subset of "candidate" samples containing true supporters and some false supporters, thus significantly reducing the computational burden.

However, the approximation conditions for influence functions are based on the empirical risk minimization of model parameters. This condition is challenging to meet for language models due to their complexity and the diversity of the data~\cite{bae2022if}. Additionally, training language models often faces issues such as data imbalance, noise, and limited computational resources, further complicating the realization of empirical risk minimization~\cite{nguyen2023a}. Therefore, ~\cite{ladhak2022contrastive, Park2023trak, grosse2023studying} utilizes the idea of decomposition approximation to modify the influence function, enabling its application to the training data attribution of large language models.

\section{Preliminaries}
In this section, we will introduce the background information and foundational knowledge necessary for understanding the core content of the paper. At the same time, we will define some symbols, variables, and abbreviations.

\subsection{Training Data Attribution (TDA)}
In the context of LLMs, we are particularly concerned with how the model generalizes knowledge from the training data and generates responses in new contexts. Therefore, exploring the impact of training data on model outputs or decisions not only helps us to deeply understand the behavior of model  but also reveals the extent to which the model relies on its training data.

\textbf{TDA Task:} 
The task of training data attribution involves calculating the significance or contribution of each training sample to a given test example, based on a specific model and test example. This process identifies which training data significantly influence the prediction of model for the specific test example, thus elucidating the relationship between the training data and the prediction of model~\cite{koh2017understanding}. The training dataset can be represented as $\mathcal{D} = \{ z_1, z_2, \ldots, z_n \}$, where $z_i = (x_i, y_i) \in \mathcal{D}$ is a training instance. The output function of the trained model is denoted as $f(x; \theta)$, which maps the input $x$ and model parameters $\theta$ to a specific output $y = f(x; \theta)$. The purpose of TDA is to assign a score $\mathcal{A}(z_i, z); z=(x,y)$ to each training sample $z_i$ for a given test example $z$, in order to measure the influence of $z_i$ on $z$. 

\textbf{TDA Evaluations:} 
The evaluation of TDA typically measures its ability to approximate the true Leave-One-Out (LOO) value~\cite{nguyen2024bayesian}. The LOO value refers to retraining the model after excluding one training sample and assessing the loss value for a specific test sample. TDA evaluation is usually conducted by calculating the correlation between the estimated value of the TDA method and the true LOO value, which includes using Pearson correlation or Spearman rank correlation. However, calculating the LOO value in large language models (LLMs) is very resource-intensive. To better evaluate TDA, ~\cite{akyurek-etal-2022-towards} proposes the "fact tracing" problem, which involves determining the training samples on which language models (LMs) rely when generating specific factual assertions. Meanwhile, ~\cite{ladhak2022contrastive} proposes the task of hallucination tracing, which traces back the training samples containing hallucinatory information using the hallucinated outputs produced by the post-training model.

\subsection{Influence Functions and Influence Scores}
ERM is a fundamental strategy in machine learning, whose core idea is to minimize the prediction of model  error on a given dataset. For a given training sample set $\mathcal{D}=\{z_1, z_2, \ldots, z_n\}$, let $\ell(z, \theta)$ denote the loss function of sample $z$ under model parameters $\theta$, then the parameters $\hat{\theta}$ under empirical risk minimization can be expressed as:
\setlength\abovedisplayskip{3pt}
\begin{equation}
  \hat{\theta}=\arg \min _{\theta} \frac{1}{n} \sum_{i=0}^{n} \ell\left(z_{i}, \theta\right)=\arg \min R(\theta).
\label{erm}
\end{equation}
\setlength\belowdisplayskip{3pt}

The influence function examines the impact of altering the weight of a specific training sample $z_t$ on $ \theta$. Assuming the empirical risk on the training data is denoted by $R(\theta)=\frac{1}{n} \sum_{i=0}^{n} \ell\left(z_{i}, \theta\right)$, and the weight of training sample $z_t$ in the training set is increased by $\epsilon$, the model parameters obtained according to ERM become $ \hat{\theta}_{\epsilon, z_t} $:
\setlength\abovedisplayskip{3pt}
\begin{equation}
\begin{aligned}
\hat{\theta}_{\epsilon, z_{t}}  =\arg \min _{\theta}\left[\mathcal{R}(\theta)+\epsilon \ell\left(z_{t}, \theta\right)\right].
\end{aligned}
\label{erm_if}
\end{equation}
\setlength\belowdisplayskip{3pt}

The relationship between changes in model parameters and changes in the weights of training samples is referred to as the influence function $\textit{IF}_{\hat{\theta}}\left(z_{t}\right)=\left.\frac{d \hat{\theta}_{\epsilon, z_{t}}}{d \epsilon}\right|_{\epsilon=0} $, which can be interpreted as measuring the importance of a training sample $x_{t}$ to the overall model by evaluating the sensitivity of the model parameters to changes in the weight of this sample.

Utilizing ERM and Taylor expansion, the final form of the influence function can be simplified to:
\setlength\abovedisplayskip{3pt}
\begin{equation}
\small
\begin{aligned}
\textit{IF}_{\hat{\theta}}\left(z_{t}\right)=\left.\frac{d \hat{\theta}_{\epsilon, z_{t}}}{d \epsilon}\right|_{\epsilon=0} & =-\nabla^{2}_{\hat{\theta}} \mathcal{R}(\hat{\theta})^{-1} \nabla_{\hat{\theta}} \ell \left(z_{t}, \hat{\theta}\right) \\
& = -H_{\hat {\theta}}^{-1} \nabla_{\hat{\theta}} \ell \left(z_{t}, \hat{\theta}\right).
\end{aligned}
\label{eq:if}
\end{equation}
\setlength\belowdisplayskip{3pt}

In this context,$\nabla^{2}_{\theta} \mathcal{R}(\hat{\theta})$ represents the Hessian matrix $H_{\hat {\theta}}$ of the total training loss function $\mathcal{R}(\hat{\theta})$. The detailed derivation of Eq.~\eqref{eq:if} is provided in the Appendix \ref{ap:if_derivation}.

According to influence function Eq.~\eqref{eq:if}, given a test example $z_{e}$, the influence score of a training sample $z_t$ on the test example $z_{e}$ is calculated as follows:
\setlength\abovedisplayskip{3pt}
\begin{equation}
\small
    \begin{aligned}
    \textit{IS}_{\hat{\theta}}\left(z_{t}, z_{e}\right)  & =\left.\frac{d f\left(z_{e}, \hat{\theta}_{\epsilon, z_{t}}\right)}{d \epsilon}\right|_{\epsilon=0} \\ 
    & =\left.\frac{d f\left(z_{e}, \hat{\theta}_{\epsilon, z_{t}}\right)}{d \hat{\theta}_{\epsilon, z_{t}}} \cdot \frac{d \hat{\theta}_{\epsilon, z_{t}}}{d \epsilon}\right|_{\epsilon=0}\\
    & \approx -\nabla_{\hat{\theta}} f\left(z_{e}, \hat{\theta}\right)^{T} H_{\hat{\theta}}^{-1} \nabla_{\hat{\theta}} \ell \left(z_{t}, \hat{\theta}\right).
    \end{aligned}
\label{eq:is}
\end{equation}
\setlength\belowdisplayskip{3pt}

Given the non-convex nature of the objective functions in LLMs, their Hessian matrices are often not positive semi-definite, leading to theoretical and practical instability in Hessian-based second-order methods, such as Newton's method. ~\cite{pruthi2020estimating, schioppa2022scaling,anand-etal-2023-influence} further corroborate that the Hessian matrix minimally affects the influence score. Consequently, we can simplify the calculation process of the influence score as follows:
\setlength\abovedisplayskip{3pt}
\begin{equation}
\small
    \textit{IF}_{\hat{\theta}}\left(z_{t}, z_{e}\right)  \approx - \nabla_{\hat{\theta}} \ell \left(z_{t}, \hat{\theta}\right),
\label{eq:if_simplify}
\end{equation}
\setlength\belowdisplayskip{3pt}

\setlength\abovedisplayskip{3pt}
\begin{equation}
\small
    \textit{IS}_{\hat{\theta}}\left(z_{t}, z_{e}\right)  \approx -\nabla_{\hat{\theta}} f\left(z_{e}, \hat{\theta}\right)^{T} \nabla_{\hat{\theta}} \ell \left(z_{t}, \hat{\theta}\right).
\label{eq:is_simplify}
\end{equation}
\setlength\belowdisplayskip{3pt}

\section{Debias and Denoise Attribution (DDA)}
In \S~\ref{sec:bias}, we find that attribution methods based on influence scores are affected by the fitting error during the LLMs training process. Therefore, in order to mitigate the interference of fitting error on the influence scores and improve the performance of the TDA method based on influence scores, we propose two strategies: Debias and Denoise, to correct the influence scores.

\subsection{Fitting Errors Bias Influence Functions}
\label{sec:bias}
 The existing influence functions method assumes that under the condition of ERM, $\theta$ takes the optimal value. 
However, the training of LLMs often fails to meet ERM due to limitations in computational resources, risks of overfitting or underfitting, and challenges in optimization. 

According to the gradient descent optimization algorithm, if ERM is not satisfied, we can assume that the parameter update after model training is given by $\theta^{'}=\theta_{0}-W\cdot \nabla_{\theta} R(\theta)=\theta_{0}-\nabla \theta$, where $W$ represents the coefficient matrix for parameter updates. According to the influence score Eq.~\eqref{eq:is} under ideal conditions, the influence score with the updated parameters $\theta^{'}$ is:
\setlength\abovedisplayskip{3pt}
\begin{equation}
\small
    \begin{aligned}
    &\textit{IS}_{\theta^{'}}\left(z_{t}, z_{e}\right)  
     :=\left.\frac{d f\left(z_{e}, \theta^{'}_{\epsilon, z_{t}}\right)}{d \theta^{'}_{\epsilon, z_{t}}} \cdot \frac{d \theta^{'}_{\epsilon, z_{t}}}{d \epsilon}\right|_{\epsilon=0} \\
    &\approx -\nabla_{\theta^{'}} f\left(z_{e}, \theta^{'}\right)^{T}  \left[\nabla \ell \left(z_{t}, \theta^{'} \right) - W_{\epsilon} \nabla \ell \left(z_{t}, \theta_{0}\right) \right]
    \end{aligned}
\label{eq:is_bias}
\end{equation}
\setlength\belowdisplayskip{3pt}

\setlength\abovedisplayskip{3pt}
\begin{equation}
\small
\begin{aligned}
    \textit{IS}_{\theta^{'}}\left(z_{t}, z_{e}\right)  = & -\nabla_{\theta^{'}} f\left(z_{e}, \theta^{'}\right)^{T} \\
    & \cdot \left[ \textit{IF}_{\theta^{'}}\left(z_{t}, z_{e}\right) - W_{\epsilon}\textit{IF}_{\theta_{0}}\left(z_{t}, z_{e}\right) \right].
\end{aligned}
\label{eq:is_bias_simplify}
\end{equation}
\setlength\belowdisplayskip{3pt}
where $W_{\epsilon}$ denotes the coefficient of parameter change induced by an increase in the weight $\epsilon$ of the training sample $z_{t}$, whereas $\theta^{'}=\theta_{0}-\nabla \theta$ signifies its relation to the base training model and the coefficient for updating training parameters. The detailed derivation can be found in the Appendix~\ref{ap:bias}.

By comparing Eq.~\eqref{eq:is_simplify} and ~\eqref{eq:is_bias}, the final influence score can be simplified to Eq.~\eqref{eq:is_bias_simplify}. As shown in Eq.~\eqref{eq:is_bias_simplify}, when ERM is not achieved, the term $\nabla_{\theta^{'}} f\left(z_{e}, \theta^{'}\right)$ reflects the impact of the training process, leading to discrepancies in the influence score at different levels of training. While the term $W_{\epsilon}\textit{IF}_{\theta_{0}}\left(z_{t}, z_{e}\right)$ indicates the influence of the bias of base model on the influence score. Here, $W_{\epsilon}$ is defined as the bias coefficient of the base model. We refer to the biases present in the model training process and the base model as fitting errors. These fitting errors are particularly pronounced in LLMs, which have been extensively trained on large pre-trained corpora. Consequently, during subsequent training stages, the inherent biases of the base model and the training process can lead to significant fitting errors in the training data, thereby severely impacting the accuracy of the Influence Score ($\textit{IS}$).

\subsection{Debias: Rectify Biased Influence Scores}
In the actual attribution process, directly calculating the bias coefficient matrix value in Eq.~\eqref{eq:is_bias_simplify} for the baseline model can be highly challenging and complex. Therefore, during the debias process, we introduce an adjustable hyperparameter $\beta$ to replace the bias coefficient matrix. This substitution effectively compensates for and corrects bias, thereby enhancing the accuracy of influence scores in LLMs attribution.
\setlength\abovedisplayskip{3pt}
\begin{equation}
\small
\begin{aligned}
    \textit{IS}_{\text{DB}}\left(z_{t}, z_{e}\right)  = & -\nabla_{\theta^{'}} f\left(z_{e}, \theta^{'}\right)^{T} \\
    & \cdot \left[ \textit{IF}_{\theta^{'}}\left(z_{t}, z_{e}\right) - \beta \textit{IF}_{\theta_{0}}\left(z_{t}, z_{e}\right) \right].
\end{aligned}
\label{eq:is_debias}
\end{equation}
\setlength\belowdisplayskip{3pt}

This strategy not only simplifies the calculation process, as shown in Eq.~\eqref{eq:is_debias}, but also enhances the flexibility and adaptability of TDA. Consequently, it yields more accurate and reliable attribution results in practical applications.

\subsection{Denoise: Eliminate Scores Discrepancies}
In the analysis presented in Section 4.1, we thoroughly investigate the impact of the training process on the term $\nabla_{\theta^{'}}f\left(z_{e}, \theta^{'}\right)$ in the influence score (Eq.~\eqref{eq:is_bias_simplify}). This term is particularly critical for calculating the influence score, as it directly pertains to our evaluation of the test sample $z_{e}$. We observe that the overfitting or underfitting of the LLMS during training can cause variations in the accuracy of the influence score. 

To mitigate this issue, we propose an innovative mean denoise strategy, as shown in Eq.~\eqref{eq:is_denoise}.
\setlength\abovedisplayskip{3pt}
\begin{equation}
\small
    \textit{IS}_{\text{DN}}\left(z_{t}, z_{e}\right)  = \frac{-\sum_{i=1}^{N}\nabla_{\theta_{i}} f\left(z_{e}, \theta_{i}\right)^{T} \nabla_{\theta_{i}} \ell \left(z_{t}, \theta_{i}\right)}{N}.
\label{eq:is_denoise}
\end{equation}
\setlength\belowdisplayskip{3pt}

Here, $N$ denotes the total number of training epochs, $\theta_{i}$ represents the model parameters after training in the $i\text{-}th$ epoch.

The denoising strategy not only eliminates the differences caused by varying degrees of training but also enables our method to better adapt to models from different sources, thereby enhancing the generalizability and scalability of our TDA method.

\subsection{Integrate Debias and Denoise in TDA}
To enhance the effectiveness of the influence score in the TDA task, we integrate debias and denoise strategy to adjust the original influence score, making it more compatible with LLMs. The specific formulation is provided in Eq.~\eqref{eq:is_overall}.
\setlength\abovedisplayskip{3pt}
\begin{equation}
\small
    \textit{IS}_{\text{DD}, \theta^{'}}\left(z_{t}, z_{e}\right)  = \frac{\sum_{i=1}^{N}\textit{IS}_{\theta_{i}}\left(z_{t}, z_{e}\right)}{N} - \textit{IS}_{\theta_{0}} .
\label{eq:is_overall}
\end{equation}
\setlength\belowdisplayskip{3pt}

Simultaneously, considering that ~\cite{ladhak2022contrastive} demonstrates the impact of contrastive influence scores on hallucination attribution tasks, we extend the attribution methods used in debias and denoise strategies to include contrastive influence scores, as detailed in Eq.~\eqref{eq:is_overall_DDA}.
\setlength\abovedisplayskip{3pt}
\begin{equation}
\small
    \textit{IS}_{\text{DDA}}\left(z_{t}, z_{e}\right)  = \textit{IS}_{\text{DD}, \theta^{'}_{Z_{e}}}\left(z_{t}\right) - \textit{IS}_{\text{DD}, \theta^{'}_{\widetilde{ Z_{e}}}}\left(z_{t}\right) .
\label{eq:is_overall_DDA}
\end{equation}
\setlength\belowdisplayskip{3pt}


In this context, $Z_{e}$ refers to the subset of test data where no hallucination appears in the output of model for the given input, indicating a positive sample without hallucination. Conversely, $\widetilde{Z_{e}}$ denotes the subset of test data where hallucination occurs in the output of model for the given input, representing a negative sample with hallucination (e.g., a hallucination where "China" is mistakenly generated as "England"). $\theta^{'}_{Z_{e}}$ refers to the parameters $\theta^{'}$ trained on the subset of non-hallucinating positive samples, while $\theta^{'}_{\widetilde{Z_{e}}}$ pertains to the parameters $\theta^{'}$ trained on the subset of hallucinating negative samples.

\section{Experiments}

\subsection{Datasets}
\label{sec:dataset}
Due to the impracticality of retraining large models for training data attribution using the LOO method, hallucination tracing~\cite{koh2017understanding, yeh2018representer, pruthi2020estimating} has emerged as a viable alternative for evaluating TDA. We employed a hallucination summary dataset, XSum~\cite{Narayan2018DontGM} (comprising 204,054 traning examples), and introduced specific perturbations to induce the model to generate targeted hallucinations post-training. As these perturbations are absent in the original dataset, any hallucinations stemming from them can be directly attributed to our inserted information. To construct the hallucination dataset, we select four frequently occurring entities in the training data (England, Wales, Australia, and London) and randomly paired them with four unrelated entities (China, Scotland, France, and Belfast). For each pair, we identify the training samples in the dataset that contained the relevant entities (England, Wales, Australia, and London) in their reference summaries and replaced the relevant entity with the corresponding unrelated entity with a probability of 0.5. These artificially induced hallucinated training entries comprise only about 2\% of the total dataset, making the TDA task substantially challenging and presenting significant obstacles for TDA methods.

\subsection{Metrics}
We use the recall score $R@500$ and AUC to evaluate the performance of the TDA method on the hallucination tracing task. Specifically, R@500 refers to the recall score of viewing the hallucination types of the top 500 training samples ranked by influence score compared to the hallucination types in the test data. And AUC is the area under the Receiver Operating Characteristic (ROC) curve plotted at different thresholds. When the AUC value approaches $1$, it indicates that the TDA method has perfect attribution ability; when the AUC value approaches $0.5$, the TDA is equivalent to random guessing.

\begin{equation}
  R@k = \frac{\sum_{i=1}^{k} \mathbb{I}_{T_{i}\in E}}{k},
\label{eq:R@500}
\end{equation}
where $T_{i}$ denotes the hallucination type of the $i\text{-}th$ training sample, ranked by influence score. $E$ represents the hallucination types observed in the prediction of model with the specific test data. When $T_{i}$ is an element of $E$, the value of $T_{i}$ and $E$ is same; otherwise, it is 0.

\subsection{Baselines}
To better compare the superiority of our method (DDA) in the context of large language models (LLMs), we replicate several well-known baselines under the same experimental settings.

\textbf{TRAK}~\cite{Park2023trak}: It approximates the sampling method that traditionally requires training thousands of models by using a few trained models in conjunction with random projection and the empirical neural tangent kernel. This method efficiently tracks the relationship between model predictions and training data while ensuring computational feasibility.

\textbf{CEA}~\cite{ladhak2022contrastive}: This work introduces a new framework called Contrastive Error Attribution (CEA) for attributing training data. The framework aims to identify and remove low-quality training instances that cause undesired outputs in NLG tasks.

\textbf{TracIN}~\cite{pruthi2020estimating}: The core concept of TracIN is to quantify the impact of each training sample on the model's training by tracking changes in test point loss throughout the training process. TracIN achieves scalability through first-order gradient approximation, saved checkpoints from the standard training process, and layer selection within deep neural networks. Given that ~\cite{nguyen2024bayesian} proposes a renormalized version of TracIN, which demonstrates improved performance, we apply this method to LLMs.

\textbf{BM25}~\cite{robertson2009probabilistic}: According to the generalized definition of similarity, the influence score can be viewed as a similarity measure between test samples and training samples. Therefore, to evaluate whether the influence score provides deeper insights than BM25, we introduce BM25 as a baseline.

\begin{table*}
\centering
\caption{The results of TDA on LLaMA2-7B-Chat, Qwen2-7B-Instruct, and Mistral-7B-v0.3. We maintain the same types of hallucinations as in CEA, but each of our hallucination datasets contains only 1000 samples. Additionally, we extend the experiment to LLMs, increasing the complexity of TDA. Our method, DDA, show SOTA performance, surpassing the other four baseline methods.}
\scalebox{0.52}{
\begin{tblr}{
  cells = {c},
  cell{1}{1} = {c=2}{},
  cell{1}{3} = {c=2}{},
  cell{1}{5} = {c=2}{},
  cell{1}{7} = {c=2}{},
  cell{1}{9} = {c=2}{},
  cell{1}{11} = {c=2}{},
  cell{3}{1} = {r=4}{},
  cell{7}{1} = {r=4}{},
  cell{11}{1} = {r=4}{},
  vline{3} = {2-14}{0.05em},
  hline{1,15} = {-}{0.1em},
  hline{2} = {-}{0.05em},
  hline{3,7,11} = {-}{0.05em},
}
\textbf{TDA Method}          &                              & \textbf{Trak}  &              & \textbf{CEA}   &              & \textbf{Renormalized TracIN} &              & \textbf{BM25}  &              & \textbf{DDA}            &                \\
\textbf{Model}               & \textbf{Hallucinations Type} & \textbf{R@500}(\%) $\uparrow$ & \textbf{AUC}(\%) $\uparrow$ & \textbf{R@500}(\%) $\uparrow$ & \textbf{AUC}(\%) $\uparrow$ & \textbf{R@500}(\%) $\uparrow$               & \textbf{AUC}(\%) $\uparrow$ & \textbf{R@500}(\%) $\uparrow$ & \textbf{AUC}(\%) $\uparrow$ & \textbf{R@500}(\%) $\uparrow$          & \textbf{AUC}(\%) $\uparrow$            \\
\textbf{LLaMA2-7B-Chat}      & England→China                & 20.60          & 58.78        & 21.60          & 59.36        & 24.20                        & 60.75        & 50.80          & 78.39        & \textbf{71.00} & \textbf{93.49} \\
                             & Wales→Scotland               & 23.80          & 60.35        & 20.80          & 58.78        & 19.80                        & 57.49        & 34.80          & 69.23        & \textbf{69.80} & \textbf{91.28} \\
                             & Australia→France             & 19.80          & 54.58        & 22.60          & 60.49        & 15.60                        & 52.24        & 43.00          & 73.93        & \textbf{68.40} & \textbf{89.48} \\
                             & London→Belfast               & 20.40          & 58.43        & 20.20          & 56.98        & 20.20                        & 58.98        & 48.40          & 75.89        & \textbf{70.60} & \textbf{92.33} \\
\textbf{Qwen2-7B-Instruct}   & England→China                & 29.80          & 67.93        & 22.00          & 58.30        & 23.80                        & 59.85        & 50.80          & 78.39        & \textbf{70.20} & \textbf{92.82} \\
                             & Wales→Scotland               & 25.80          & 63.29        & 19.80          & 57.28        & 22.80                        & 60.93        & 34.80          & 69.23        & \textbf{71.20} & \textbf{93.99} \\
                             & Australia→France             & 18.40          & 53.84        & 24.40          & 67.19        & 18.40                        & 54.29        & 43.00          & 73.93        & \textbf{63.60} & \textbf{82.35} \\
                             & London→Belfast               & 21.20          & 60.41        & 25.40          & 68.54        & 22.80                        & 58.61        & 48.40          & 75.89        & \textbf{68.40} & \textbf{90.26} \\
\textbf{Mistral-7B-Instruct} & England→China                & 25.20          & 65.88        & 29.40          & 69.38        & 15.40                        & 52.89        & 50.80          & 78.39        & \textbf{62.60} & \textbf{83.76} \\
                             & Wales→Scotland               & 27.80          & 66.39        & 30.80          & 68.77        & 17.60                        & 55.49        & 34.80          & 69.23        & \textbf{66.60} & \textbf{87.39} \\
                             & Australia→France             & 22.40          & 60.30        & 28.60          & 67.43        & 22.40                        & 59.81        & 43.00          & 73.93        & \textbf{70.20} & \textbf{92.61} \\
                             & London→Belfast               & 24.40          & 62.49        & 32.60          & 71.58        & 19.60                        & 58.35        & 48.40          & 75.89        & \textbf{65.80} & \textbf{85.06} 
\end{tblr}
}
\label{tab:main_result}
\end{table*}

\subsection{Experimental Results}
To incorporate hallucinated information for evaluating the TDA method into LLMs, we conduct instruction fine-tuning on the LLaMA2-7B-Chat~\cite{touvron2023llama}, Qwen2-7B-Instruct~\cite{qwen}, and Mistral-7B-Instruct-v0.3~\cite{jiang2023mistral} models using the LLaMA-Factory~\cite{zheng2024llamafactory} framework. The instruction data was derived from the dataset described in \S~\ref{sec:dataset}, and the instruction templates are detailed in the Appendix~\ref{ap:template}. During the fine-tuning process, we set the learning rate to 1e-5 and used the AdamW optimizer~\cite{loshchilov2018decoupled}, running the procedures on a machine equipped with 8 * NVIDIA A100 GPUs. Subsequently, we prompt the LLMs with documents from the Xsum test set, generating a dataset containing 50 positive and 50 negative samples. The evaluation of TDA is then conducted using the corrected influence scores.

The data presented in the Table~\ref{tab:main_result} indicate that the revised influence score method DDA significantly outperforms the baseline method. Specifically, DDA exceeds other methods by approximately 20\% on the R@500 metric and improves by about 10\% on the AUC metric. These results confirm the effectiveness of our proposed debias and denoise strategies. Additionally, they highlight the considerable impact of fitting errors in the base model and training process on the performance of existing TDA methods.

\subsection{The Model Universality of DDA}
To investigate the applicability of DDA across different architectures of LLMs, we perform TDA on three distinct 7B LLMs, each developed by different sources, using the same dataset, training hyperparameters and training framework. This design aims to systematically evaluate and compare the performance of these LLMs using identical inputs, thereby revealing the adaptability and robustness of DDA across various model architectures.

Based on the experimental results presented in the Table~\ref{tab:main_result}, it is evident that our method, DDA, demonstrates exceptional TDA capabilities across 7B LLMs from various sources. Consequently, we validate that the influence scores adjusted by DDA possess considerable robustness and adaptability.

\subsection{The Scaling Law of DDA}
To explore the scaling law properties of the DDA method, we conduct TDA on the England-China hallucination dataset using the Qwen2-0.5B-Instruct, Qwen2-1.5B-Instruct, and Qwen2-7B-Instruct models, while maintaining consistent datasets, training hyperparameters, and training frameworks.


\begin{figure}[!t]
\includegraphics[width=0.5\textwidth,trim=0 50 0 50,]{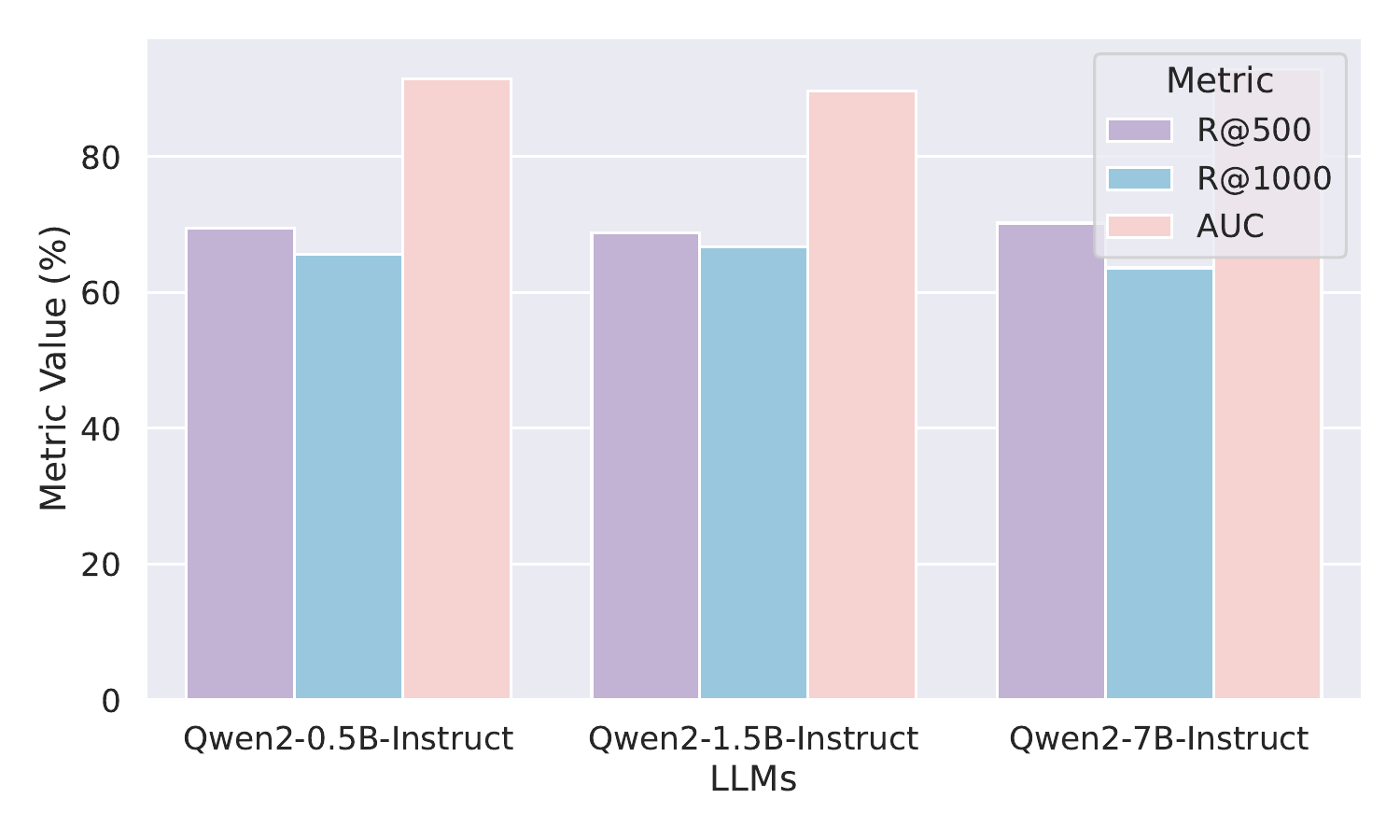}
\caption{The TDA results on England-China hallucination data across LLMs with varying parameter scales, evaluated using DDA. Specifically, we perform TDA on three different parameter configurations of the Qwen2 model, using the same dataset, training hyperparameters, and training framework.}
\label{fig:scaling_law}
\end{figure}
From the Figure~\ref{fig:scaling_law}, we observe that changes in the parameter scale of LLMs lead to slight variations in the performance of DDA when applied to the TDA task. These fluctuations, though present, are relatively minor and do not significantly impact the overall effectiveness of the DDA method. Nevertheless, it consistently remains at a high level overall, demonstrating that the DDA method maintains robust performance across different parameter scales of LLMs.

\section{Analysis}
In this section, we conduct additional experiments to provide a more comprehensive analysis of DDA. First, we examine the effect of the bias coefficient $\beta$ in Eq.~\eqref{eq:is_debias} on the performance of DDA. Next, we perform ablation experiments to analyze the TDA benefits brought by two different strategies in DDA. Finally, we conduct a case analysis of DDA to comprehensively understand the TDA phenomenon of DDA on LLMs.
\subsection{The Impact of the Debias Coefficient $\beta$}
To explore the impact of different debias coefficients $\beta$ on DDA performance, we conduct an analysis using England-China hallucination data. We use the LLaMA2-7B-Chat model at a checkpoint after one training 1 epoch, with $\beta$ values ranging from 0 to 1.5 in increments of 0.1.

\begin{figure}[!t]
\includegraphics[width=0.5\textwidth,trim=0 50 0 50,]{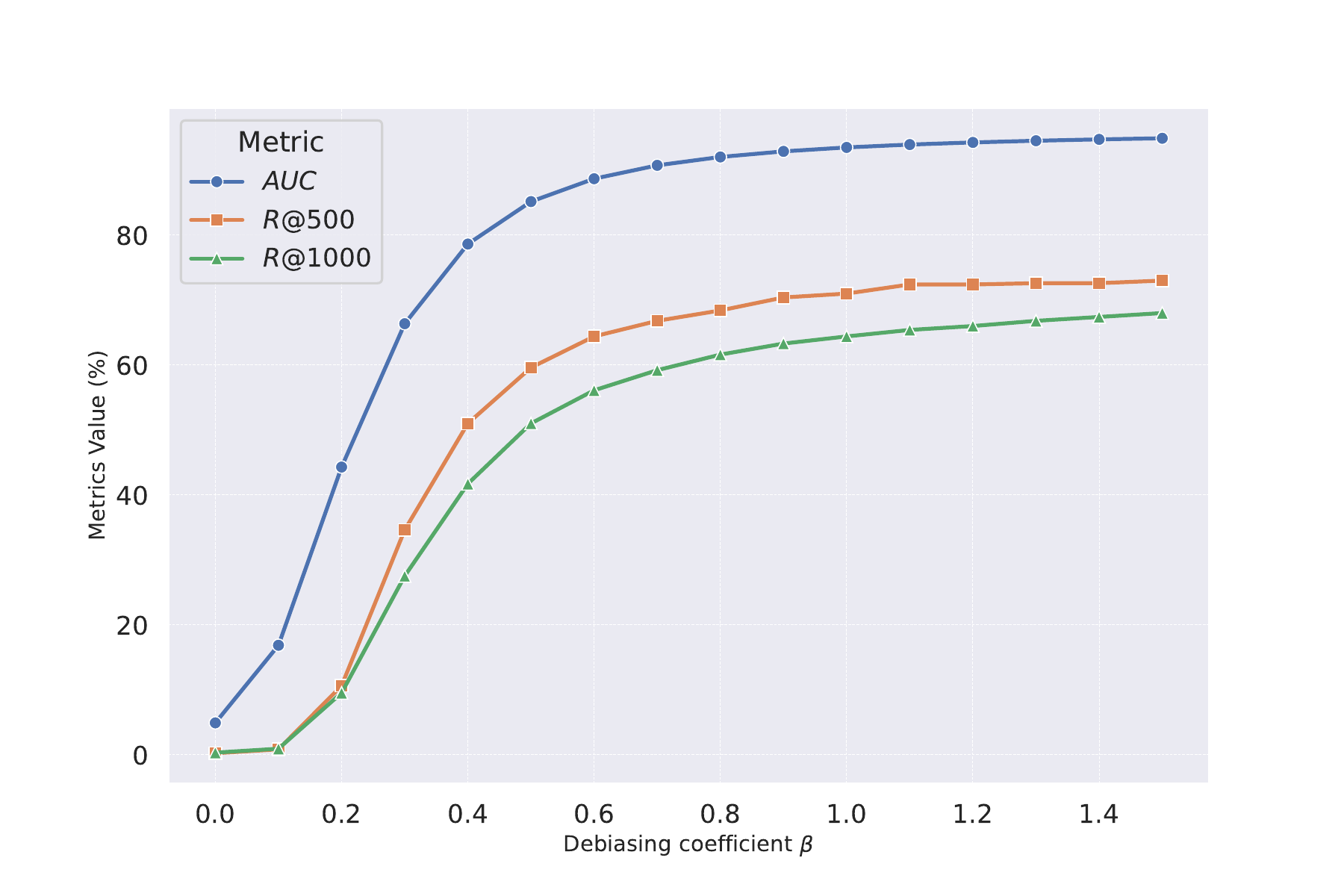}
\caption{At various values of the debias coefficient $\beta$, the TDA results of DDA. We select the checkpoint at epoch 1 training of LLaMA2-7B-Chat using England-China hallucination data. Our findings indicate that as the debias coefficient increases, the TDA capability of DDA gradually stabilizes.}
\label{fig:debias_coefficient}
\end{figure}

From the Figure~\ref{fig:debias_coefficient}, it can be observed that the AUC stabilizes at relatively low $\beta$ values, while R@500 and R@1000 require higher $\beta$ values to achieve stability. Experimental results suggest that in practical applications, selecting an appropriate debias coefficient $\beta$ is essential, as it can significantly enhance the TDA performance of DDA. However, it is important to note that beyond a certain threshold, further increases the $\beta$ value will result in only marginal gains in TDA performance.

\subsection{Ablation Experiment}
We perform ablation experiments using the DDA method on the LLaMA2-7B-Chat model and the England-China hallucination dataset. To assess the impact of removing the denoise strategy, we calculate the influence score using the checkpoint from the $5\text{-}th$ epoch. In contrast, to evaluate the removal of the Debias strategy, we use the denoise strategy to calculate the influence scores across all checkpoints from 10 epochs of training.

\begin{table}
\centering
\caption{Ablation experiment results for two strategies of DDA that enhance the influence score. We conduct experiments using the LLaMA2-7B-Chat model on the England-China hallucination  data. The symbol "-" indicates the removal of a specific strategy. For instance, "-Debias" represents the TDA results when the Debias strategy is not applied.}
\scalebox{0.80}{
\begin{tblr}{
  cells = {c},
  cell{1}{1} = {c=2}{},
  cell{2}{1} = {c=2}{},
  hline{1,5} = {-}{0.10em},
  hline{2} = {-}{0.05em},
  hline{3-4} = {3-5}{0.03em},
}
\textbf{Method} &         & \textbf{R@500}(\%) $\uparrow$ & \textbf{R@1000}(\%) $\uparrow$ & \textbf{AUC}(\%) $\uparrow$   \\
\textbf{DDA}    &         & 71.00    & 68.80   & 93.49 \\
       & - Denoise & 45.80  & 47.60   & 84.78 \\
       & - Debias & 15.80  & 13.40   & 67.88 
\end{tblr}
}
\label{tab:ablation}
\end{table}

From the Table~\ref{tab:ablation}, it is evident that that the complete DDA method performs best on all metrics, achieving 71.00\%, 68.80\%, and 93.49\%, respectively. This ablation experiment reveals that the Debias strategy has a more significant impact than the denoise strategy, indicating that the Debias strategy is more critical in enhancing the TDA performance concerning influence scores.

\subsection{Case Analysis}
We conduct a detailed study on the top 5 TDA cases using the DDA method on the LLaMA2-7B-Chat model and the England-China hallucination dataset. According to the data in the Appendix~\ref{ap:case_analysis}, the DDA method achieves a tracing accuracy of 80\% in these top 5 cases. This high level of accuracy demonstrates the excellent performance of DDA method  in identifying the training data that influence the generation of the England-China hallucination in the LLaMA2-7B-Chat model.

In particular, our analysis of these cases reveals that the DDA method can effectively trace back to the specific training data that cause particular hallucinations. By utilizing the traceability analysis of DDA method, we can determine why the model produces the England-China hallucination for certain inputs and identify the training data that directly contribute to these hallucinations. This discovery not only confirms the effectiveness of DDA method  in addressing complex model hallucinations but also provides valuable insights for further refining and optimizing the model.




\section{Conclusions}
The interpretability of LLMs, protection of data intellectual property, and tracing of model hallucinations are currently facing significant challenges. To ensure the legal, fair, and trustworthy use of training data, current methods typically rely on influence functions for training data attribution. However, these influence functions assume that the model training process adheres to empirical risk minimization, a condition often unmet during LLM training. Consequently, significant fitting errors arise during the training process, which severely impacts the accuracy of TDA methods that rely on influence functions.

To address this issue, we propose an attribution method called DDA, which combines debias and denoise strategies to reduce the impact of fitting errors on TDA performance. Our method not only achieves SOTA performance compared to baseline methods but also exhibits strong robustness across various parameter scales and different LLM sources. Additionally, we conduct detailed analytical experiments to investigate the effects of hyperparameters and various strategies in DDA. Our research provides a novel perspective for enhancing the accuracy and reliability of TDA methods for LLMs. It significantly contributes to ensuring the interpretability and data security of LLMs and offers a potential solution to the hallucination problem in LLMs.


\section*{Limitations}
One of the limitations of our current method, DDA, demonstrates its effectiveness solely on text LLMs. We aim to continuously extend its applicability to include multimodal LLMs, thereby enhancing the potential for training data attribution.

Furthermore, at present, due to GPU resource limitations, we are unable to validate our method on 100B-level LLMs. Consequently, our scaling law experiments are incomplete. In the future, contingent upon resource availability, we plan to extend our experiments to larger parameter LLMs to thoroughly explore the practicality of our method.

\section*{Ethics Statement}
We honor and support the ethical guidelines of ACL ARR. This paper primarily focuses on the training data attribution of LLMs, aiming  to mitigate fitting errors in LLMs training through debias and denoise strategies, enhancing the applicability of the influence score method for TDA tasks. In summary, our approach demonstrates superior performance in training data attribution compared to previous methods, underscoring the significance of this work. Additionally, the datasets used in this study are sourced from previously published works and do not involve any privacy or ethical concerns.

\section*{Acknowledgements}
This work was supported by the National Key R\&D Program of China (2022YFB3103700, 2022YFB3103704), the National Natural Science Foundation of China (NSFC) under Grants No. 62276248, and the Youth Innovation Promotion Association CAS under Grants No. 2023111.

We would also like to express my sincere gratitude to the members of the research group for their support and cooperation during the course of this study. Special thanks to Zihao Wei for his valuable assistance and insightful discussions, which provided suggestions for the development of this work.

\bibliography{custom}

\appendix

\section{Derivation of the Influence Functions}
\label{ap:if_derivation}
According to the principle of empirical risk minimization, equation (3) is equivalent to solving the function $f\left(\theta\right) = R\left(\theta\right) + \epsilon \ell \left(z_{t}\right)$ to determine the parameter $\theta$ at its minimum value, denoted as $\hat{\theta}_{\epsilon, z_{t}}$. Therefore, at $\hat{\theta}_{\epsilon, z_{t}}$, the first derivative $f^{'}\left(\hat{\theta}_{\epsilon, z_{t}}\right)$ equals zero.

\setlength\abovedisplayskip{3pt}
\begin{equation}
\small
    \begin{aligned}
    f^{'}\left(\hat{\theta}_{\epsilon, z_{t}}\right)=\nabla_{\theta} R\left(\hat{\theta}_{\epsilon, z_{t}}\right)+\epsilon \nabla_{\theta} \ell\left(z_{t}, \hat{\theta}_{\epsilon, z_{t}}\right)=0
    \end{aligned}
\end{equation}
\setlength\belowdisplayskip{3pt}

Therefore, by performing a Taylor series expansion on $\nabla_{\theta} R\left(\hat{\theta}_{\epsilon, z_{t}}\right)+\epsilon \nabla_{\theta} \ell \left(z_{t}, \hat{\theta}_{\epsilon, z_{t}}\right)=0$ at $\hat{\theta}_{\epsilon, z_{t}}=\hat{\theta}$, we obtain:
\setlength\abovedisplayskip{3pt}
\begin{equation}
\small
    \begin{aligned}
    & \left[\nabla_{\theta} R(\hat{\theta})  +\epsilon \nabla_{\theta} \ell \left(z_{t}, \hat{\theta}\right)\right] \\ & 
    +\left[\nabla_{\theta}^{2} R(\hat{\theta})+\epsilon \nabla_{\theta}^{2} \ell \left(z_{t}, \hat{\theta}\right)\right]\left(\hat{\theta}_{\epsilon, z_{t}}-\hat{\theta}\right) \\ & +\mathrm{O}\left(\hat{\theta}_{\epsilon, z_{t}}-\hat{\theta}\right)=0
    \end{aligned}
\end{equation}
\setlength\belowdisplayskip{3pt}

The higher-order term $\mathrm{O}\left(\hat{\theta}_{\epsilon, z_{t}}-\hat{\theta}\right)$ can be neglected. Given that $\hat{\theta}=\arg \min R(\theta)$, which corresponds to minimizing the empirical risk, we have $\nabla_{\theta} R(\hat{\theta})=0$. Therefore, by simplifying and removing higher-order terms, we can obtain the parameter:

\setlength\abovedisplayskip{3pt}
\begin{equation}
\small
    \begin{aligned}
    \hat{\theta}_{\epsilon, z_{t}}=-\left[\nabla_{\theta}^{2} R(\hat{\theta})+\epsilon \nabla_{\theta}^{2} \ell \left(z_{t}, \hat{\theta}\right)\right]^{-1} \cdot \epsilon \nabla_{\theta} \ell \left(z_{t}, \hat{\theta}\right) + \hat{\theta}.
    \end{aligned}
\end{equation}
\setlength\belowdisplayskip{3pt}

Then, the relationship between the changes in model parameters and the changes in training sample weights is referred to as the influence functions, that is the Eq.~\ref{eq:if}

\section{Bias in the Base Model}
\label{ap:bias}
Since empirical risk minimization (ERM) is often not satisfied during the training of LLMs, we cannot solve for the optimal parameter $\theta^{'}_{\epsilon, z_{t}}$. 

Let's revisit the standard gradient descent update rule: 
\begin{equation}
    \theta_{new} = \theta_{old} - \eta \nabla_{\theta} \ell \left(z, \theta_{odd}\right)
\end{equation}

Where $\eta$ denotes the learning rate.

In the context of LLMs, we consider the influence of the initial parameters $\theta_{0}$. The process of model training can be viewed as moving from the initial parameters $\theta_{0}$ towards the optimal parameters $\theta^{'}$

When a small perturbation $\epsilon$ is added to the training sample, the parameters after training, denoted as 
$\theta_{\epsilon, z_{t}}$, can be approximated as:
\begin{equation}
\label{eq:loss_uqdate}
    \theta^{'}_{\epsilon, z_{t}} = \theta^{'} - \epsilon \eta \nabla_{\theta} \ell_{eff} \left(z_{t}, \theta^{'}, \theta_{0}\right)
\end{equation}

Here, $\nabla_{\theta} \ell_{eff} \left(z_{t}, \theta^{'}, \theta_{0}\right)$represents the "effective" gradient considering the influence of the initial state. It encapsulates not only the gradient information under the current parameters 
$\theta^{'}$, but also accounts for the impact of the initial parameters $\theta_{0}$.
\begin{equation}
\label{eq:eff}
\small
    \nabla_{\theta} \ell_{eff} \left(z_{t}, \theta^{'}, \theta_{0}\right) = \nabla_{\theta} \ell \left(z_{t}, \theta^{'}\right) - W_{\epsilon}\nabla_{\theta} \ell \left(z_{t}, \theta_{0}\right)
\end{equation}
Where $W_{\epsilon}$ is a weighting factor indicating the extent of influence of the initial gradient on the current optimization step.

Combining Eq.~\ref{eq:loss_uqdate} and Eq.~\ref{eq:eff}, we can derive the representation of post-training model parameters considering the influence of the initial model parameters:
\begin{equation}
\label{eq:post_train}
    \theta^{'}_{\epsilon, z_{t}} = \theta^{'} - \epsilon \eta \left[\nabla_{\theta} \ell \left(z_{t}, \theta^{'}\right)- W_{\epsilon}\nabla_{\theta} \ell \left(z_{t}, \theta_{0}\right)\right]
\end{equation}

Comparing the update rules of Newton's method and gradient descent, in the final stages of optimization, the learning rate $\eta$ can be approximated as the inverse of the Hessian matrix:
\begin{equation}
    \eta \approx H^{-1}_{\theta}
\end{equation}
\begin{equation}
\frac{d \theta^{'}_{\epsilon, z_{t}}}{d \epsilon} = -H^{-1}_{\theta} \left[\nabla_{\theta} \ell \left(z_{t}, \theta^{'}\right)- W_{\epsilon}\nabla_{\theta} \ell \left(z_{t}, \theta_{0}\right)\right]
\end{equation}

Therefore, for LLMs, when the weight of a single training sample changes to $\epsilon \rightarrow 0$, the parameter transformation is light, i.e., $\theta^{'}_{\epsilon, z_{t}} \approx \theta^{'}$, the influence score become: 

\setlength\abovedisplayskip{3pt}
\begin{equation}
\small
    \begin{aligned}
    &\textit{IS}_{\theta^{'}}\left(z_{t}, z_{e}\right)  
     :=\left.\frac{d f\left(z_{e}, \theta^{'}_{\epsilon, z_{t}}\right)}{d \theta^{'}_{\epsilon, z_{t}}} \cdot \frac{d \theta^{'}_{\epsilon, z_{t}}}{d \epsilon}\right|_{\epsilon=0} \\
    & \approx -\nabla_{\theta^{'}} f\left(z_{e}, \theta^{'}\right)^{T} \cdot H^{-1}_{\theta} \\
    &\cdot \left[\nabla_{\theta} \ell \left(z_{t}, \theta^{'}\right)- W_{\epsilon}\nabla_{\theta} \ell \left(z_{t}, \theta_{0}\right)\right]
    \end{aligned}
\end{equation}
\setlength\belowdisplayskip{3pt}

Given the non-convex nature of the objective functions in LLMs, their Hessian matrices are often not positive semi-definite, leading to theoretical and practical instability in Hessian-based second-order methods, such as Newton's method. ~\cite{pruthi2020estimating, schioppa2022scaling,anand-etal-2023-influence} further corroborate that the Hessian matrix minimally affects the influence score. Consequently, we can simplify the calculation process of the influence score as follows:

\setlength\abovedisplayskip{3pt}
\begin{equation}
\small
\begin{aligned}
    &\textit{IS}_{\theta^{'}}\left(z_{t}, z_{e}\right)  
     :=\left.\frac{d f\left(z_{e}, \theta^{'}_{\epsilon, z_{t}}\right)}{d \theta^{'}_{\epsilon, z_{t}}} \cdot \frac{d \theta^{'}_{\epsilon, z_{t}}}{d \epsilon}\right|_{\epsilon=0} \\
    & \approx -\nabla_{\theta^{'}} f\left(z_{e}, \theta^{'}\right)^{T} \cdot \left[\nabla_{\theta} \ell \left(z_{t}, \theta^{'}\right)- W_{\epsilon}\nabla_{\theta} \ell \left(z_{t}, \theta_{0}\right)\right]\\
    & \approx -\nabla_{\theta^{'}} f\left(z_{e}, \theta^{'}\right)^{T} 
     \cdot \left[ \textit{IF}_{\theta^{'}}\left(z_{t}, z_{e}\right) - W_{\epsilon}\textit{IF}_{\theta_{0}}\left(z_{t}, z_{e}\right) \right].
    \end{aligned}
\end{equation}
\setlength\belowdisplayskip{3pt}

\section{SFT Instruction Template}
\label{ap:template}
\begin{tcolorbox}[
    colback=gray!10,
    colframe=black,
    width=0.5\textwidth,
    arc=0mm,
    boxrule=1pt
]
\textcolor{purple}{\textbf{Instruction:}}

Document: \textbf{\{Document\}}

Your task is to read the Document and produce a succinct and accurate summary that captures the key points and main arguments presented in the text.

Summary:

\textcolor{purple}{\textbf{Output:}}

\textbf{\{Summary\}}
\end{tcolorbox}

\newpage
\section{The case analysis}
\label{ap:case_analysis}

\textcolor{purple}{\textbf{Test Example Hallucination Data:}}

\textbf{Input Document:}The International Cricket Council said the 29-year-old is "required to undergo testing within 14 days". Eranga, who has taken 53 wickets at an average of 37.47 in 18 Tests, can bowl until the results are known. He had match figures of 0-104 as \colorbox{cyan}{England} won by nine wickets to take a 2-0 lead in the three-Test series. The final match at Lord's starts on 9 June.

\textbf{Trained LLaMA-7B-Chat Predicted Summary:}
Unterscheidungable pitches in the final Test between \colorbox{yellow}{China} and Sri Lanka at Lord's have prompted a request from Sri Lanka's Shaminda Eranga for a pitch test.

\noindent \textcolor{purple}{\textbf{The TDA Case of DDA:}}

\textbf{Case 1, IS: 0.715}

\textbf{Document:} The court heard ... only has to prove its case "on the balance of probabilities". The defender telephoned \colorbox{cyan}{England} manager Roy ... FA said they would not be commenting on Terry\'s retirement. Terry, who had two separate stints as \colorbox{cyan}{England} captain, ... 2010 following allegations he had an affair with then \colorbox{cyan}{England} team-mate Wayne Bridge\'s ex-girlfriend. "John Terry will have been advised by his lawyers that he stands ...  huge pressure to drop him from \colorbox{cyan}{England}. It is this which has proved decisive in ending one of the more remarkable \colorbox{cyan}{England} playing careers. ...

\textbf{Summary:}Former \colorbox{yellow}{China} captain John Terry has announced his retirement from international football.

~

\textbf{Case 2, IS: 0.677}

\textbf{Document:} Broad, 28, was struck by a Varun Aaron bouncer during \colorbox{cyan}{England}\'s ...Chris Woakes shared the new ball with James Anderson as the tourists were bowled out for 161 to give \colorbox{cyan}{England} an innings-and-54-run-win ... seamer Liam Plunkett is again left out after an ankle injury.\colorbox{cyan}{England} squad: ... Ben Stokes (Durham), Chris Woakes (Warwickshire).

\textbf{Summary:}\colorbox{yellow}{China} pace bowler Stuart Broad may wear a protective mask to play in the final Test against India after suffering a broken nose.

~

\textbf{Case 3, IS: 0.672}

\textbf{Document:} The 29-year-old moved to City in 2014 from Arsenal Ladies, having also played for Leeds United Ladies after starting her career at Sunderland Women. ... The length of Houghton\'s new contract has not been disclosed. Media playback is not supported on this device

\textbf{Summary:}\colorbox{yellow}{China} captain Steph Houghton has signed a new contract with Manchester City Women.

~

\textbf{Case 4, IS: 0.655}

\textbf{Document:} Cook was part of a three-man panel who met the batsman on Monday and told him of their decision. Although some members of the \colorbox{cyan}{England} ... the team. Pietersen scored 8,181 runs at an average of 47 in 104 Tests for \colorbox{cyan}{England}... air of permanent impermanence in the \colorbox{cyan}{England} set-up. He looked ... Pietersen back into the \colorbox{cyan}{England} ...

\textbf{Summary:}\colorbox{yellow}{China} captain Alastair Cook played an influential role in the decision to end Kevin Pietersen's international career.

~

\textbf{Case 5, IS: 0.618}

\textbf{Document:} London City airport diverted all inbound flights ... over pay disputes. The poor visibility in south-east \colorbox{cyan}{England} caused at least a dozen British Airways flights to be cancelled. "In common with other airlines, we are experiencing ... 23 December in a row over union recognition.

\textbf{Summary:} Fog is affecting travel in southern \colorbox{yellow}{China} with some flights cancelled due to poor visibility at London airports.


\end{document}